\pdfoutput=1

\documentclass[11pt]{article}

\usepackage{acl}
\usepackage{times}
\usepackage{latexsym}
\usepackage{graphicx}
\usepackage{amsmath}
\usepackage{parskip}
\usepackage{colortbl}
\usepackage{tikz}
\usepackage{float}
\usepackage{placeins}
\newcommand{\autour}[1]{\tikz[baseline=(X.base)]\node [draw=white,fill=cyan!20,thick,rectangle,inner sep=3pt, rounded corners=4pt] (X) {#1};}
\newcommand{\redtour}[1]{\tikz[baseline=(X.base)]\node [draw=white,fill=red!20,thick,rectangle,inner sep=3pt, rounded corners=4pt] (X) {#1};}

\newcommand{\thickhline}{%
    \noalign {\ifnum 0=`}\fi \hrule height 1pt
    \futurelet \reserved@a 
}

\usepackage{times}
\usepackage{latexsym}

\usepackage[T1]{fontenc}

\usepackage[utf8]{inputenc}

\usepackage{microtype}

\title{Detoxifying Language Models with a Toxic Corpus}

 \author{Yoon A Park \textsuperscript{1,2},  Frank Rudzicz  \textsuperscript{1,2,3}\\
  \textsuperscript{1} University of Toronto, 
  \textsuperscript{2} Vector Institute of Artificial Intelligence, 
  \textsuperscript{3} Unity Health Toronto \\ 
  \texttt{\{ypark, frank\}@cs.toronto.edu}} 

\graphicspath{{Images/}}
\begin{document}
\maketitle
\begin{abstract}
Existing studies have investigated the tendency of autoregressive language models to generate contexts that exhibit undesired biases and toxicity. Various debiasing approaches have been proposed, which are primarily categorized into data-based and decoding-based. In our study, we investigate the ensemble of the two debiasing paradigms, proposing to use toxic corpus as an additional resource to reduce the toxicity. Our result shows that toxic corpus can indeed help to reduce the toxicity of the language generation process substantially, complementing the existing debiasing methods. 

\end{abstract}

\section{Introduction}
Pretraining language models (LMs) have been a foundation of NLP given recent performance achievements; however, there is a growing concern related to inherent societal and harmful biases in these models. Due to historical biases embedded in training corpora, it is unavoidable for the language models to absorb, reproduce, and even amplify such undesired biases \citep{Schick2020Self-DiagnosisNLP}. 

\citet{Gehman2020REALTOXICITYPROMPTS:Models} showed that pretrained LMs generate toxic text even when conditioned on innocuous prompts. One of their proposed debiased techniques is Domain-Adaptive Pretraining \citet{Gururangan2020DontTasks}, or DAPT, on a non-toxic corpus. \citet{Schick2020Self-DiagnosisNLP} proposed a self-debiasing approach that uses only a handful of templates that contain the definition of undesired attributes. DAPT is a data-based approach where internal weights are updated with an additional phase of pretraining. On the other hand, self-debiasing is a decoding-based approach that does not require additional resources. The difference between the two debiasing paradigms is a trade-off between the computational cost and the quality of debiasing.

In this study, we propose to ensemble the data- and decoding-based approaches by using a toxic corpus as a detoxifying strategy. Our study attempts to invalidate the belief that only non-toxic corpora can reduce the toxicity of language generation. We use  GPT-2 \citep{Radford2018LanguageLearners} as our primary language model and OpenWebText (OWTC; \citealp{Gokaslan2019OpenWeb}), a large corpus of English webtext, as our training corpus. We measure the toxicity of each document using PerspectiveAPI\footnote{https://www.perspectiveapi.com/} and collect non-toxic and toxic corpora that satisfy our toxicity requirements. 

Our results demonstrate that using the toxic corpus indeed reduces the toxicity level of text generated from pretrained language models, which can be further improved by ensemble with the non-toxic corpus.


\section{Background and Related Work}

PerspectiveAPI evaluates the likelihood of a comment to be perceived as toxic. It divides the toxicity into eight emotional attributes, including toxicity, severe toxicity, identity attack, insult, threat, profanity, sexual explicit, and flirtation. The model is a multilingual BERT-based model, distilled into a single-language convolutional neural network (CNN).  
The AUC of the model on test sets ranges between 0.97 to 0.99 \footnote{\label{auc_link}https://developers.perspectiveapi.com/s/about-the-api-best-practices-risks}, which we safely assume to use to classify the documents. 

The model is also evaluated on the bias across a range of identity terms. Test sets are generated by swapping the identity terms on both toxic and non-toxic sentences. In English test sets, the AUC of all the identity terms fall between 0.96 to 1.0, which indicates unbiased evaluation across the different identity groups. 

\subsection{Bias in NLP}
Language embeddings or LMs are prone to unintended biases against the under-represented minority groups and inherent toxicity \citep{10.5555/3157382.3157584, manzini2019black}. Contextualized embeddings like ELMo and BERT have also proven to inherit biases, such as gender bias \citep{zhao-etal-2019-gender, zhao-etal-2018-gender}. Language generation also suffers from varying types of social biases such as stereotypical bias \citep{liang2021understanding} and sentiment bias \citep{huang2020reducing}. 

Along with the detection of bias in language embeddings and models, various fairness benchmarking \citep{Nangia2020CrowS-Pairs:Models, Dhamala2021BOLD:Generation} and debiasing approaches have been proposed. \citet{10.5555/3157382.3157584} and \citet{Liang2020TowardsRepresentations} proposed to find the hypothetical bias dimension in embedding spaces. \citet{Liu2020MitigatingLearning} proposed adversarial learning to disentangle biased and unbiased features in dialogue systems. While most of the work in fairness in NLP focuses on stereotypical biases, other studies focus on the toxicity of LMs \citep{Gehman2020REALTOXICITYPROMPTS:Models, welbl2021challenges, Schick2020Self-DiagnosisNLP}, which are most relevant to our study.

\subsection{Toxicity of Autoregressive Language Models and Debiasing}
Autoregressive pretrained language models suffer from unintended toxicity. \citet{Gehman2020REALTOXICITYPROMPTS:Models} demonstrated that the majority of pretrained models generate toxic context and investigated various detoxifying strategies. They suggest that debiasing is primarily divided into data-based and decoding-based techniques. Data-based techniques involve additional pretraining, such as domain-adaptive pretraining \citep{Gururangan2020DontTasks}, attribute conditioned pretraining, and PPLM \citep{Dathathri2020PLUGGENERATION}. These are effective but costly due to multiphase pretraining. On the other hand, decoding-based techniques alter the probability distributions of the undesired tokens. Examples include word filtering, vocabulary shifting \citep{Ghosh2017Affect-LM:Generation}, and self-debiasing \citep{Schick2020Self-DiagnosisNLP}. Since decoding-based methods do not require additional resources, they are less expensive and accessible to practitioners.

According to \citet{Gehman2020REALTOXICITYPROMPTS:Models}, adapting pretraining on non-toxic corpus is one of the effective debiasing methods despite its simplicity. In our study, we investigate whether a toxic corpus, combined with a decay function (eq.~\ref{eq:1}), can further detoxify the language generation process.



\section{Experimental Setup}

\begin{figure}[htp]
    \centering
    \includegraphics[width=7cm]{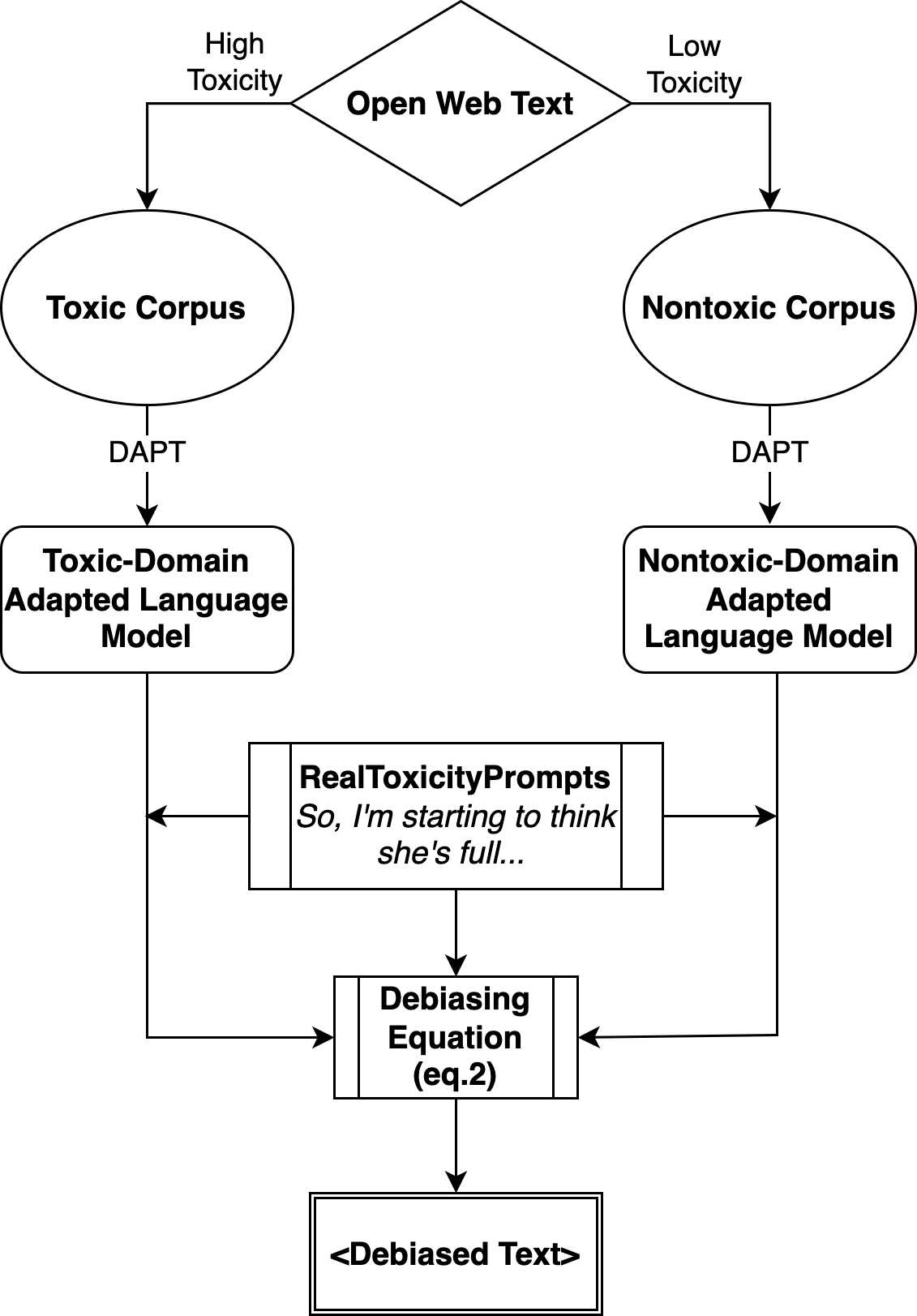}
    \caption{A flowchart of the pipeline that ensembles the data-based and decoding-based approach using both toxic and non-toxic corpus.}
    \label{fig:flowchart}
\end{figure}

\subsection{Prompts Dataset} 
\citet{Gehman2020REALTOXICITYPROMPTS:Models} released RealToxicityPrompts to compare the toxicity of conditional language generation among various LMs. Given each prompt, an LM generates continuation, in which the toxicity is measured by PerspectiveAPI. In our experiment, we use 1,225 prompts categorized as "challenging", since all out-of-the-shelf LMs tested by \citet{Gehman2020REALTOXICITYPROMPTS:Models} generated toxic sentences conditioned on these prompts. 

In addition to the RealToxicityPrompts dataset, we test our debiasing methods on the BOLD dataset \citep{Dhamala2021BOLD:Generation}, a bias benchmarking dataset covering five domains -- gender, race, political ideology, religious ideology, and profession. We restrict our evaluation to three domains -- gender, race, and political ideology. 

\FloatBarrier
\begin{table*}
    \centering
    \begin{tabular}{lcc|cc|c}
 
\textbf{Corpus} & \multicolumn{2}{c|}{\textbf{Non-Toxic}} & \multicolumn{2}{c|}{\textbf{Toxic}} & \textbf{All}\\ 
\hline
 \textbf{Percentile} & $\leq 2$&  $\leq 5$   & $\geq 95$  & $\geq 98$  &  \\
 \textbf{Avg Toxicity} & 1.42 (\%) & 2.44 (\%)  & 55.9 (\%)& 65.8 (\%)& 15.7 (\%)\\
 \textbf{Data Size} & 290 MB & 722 MB  & 981 MB & 376 MB & 16.8 GB\\
    \end{tabular}
    \caption{Average toxicity of OpenWebText by percentile.}
    \label{tab:corpus}
\end{table*}
\FloatBarrier

\subsection{Toxic Corpus Creation}  \label{sec:corpus_creation}
We use OpenWebText (OWTC; \citealp{Gokaslan2019OpenWeb}) to extract a target corpus for adaptive pretraining. OWTC is an open-source replica of OPENAI WebText \citep{Radford2018LanguageLearners}, a training corpus for GPT-2. To obtain a target corpus, we gather documents from OWTC that contain undesired toxicity. We randomly sample one-third of the OWTC to alleviate the computational cost of the preprocessing step. Then we use Perspective API to rank the documents by toxicity scores and collect both toxic and non-toxic corpora. At the end of preprocessing, we have four target corpora, two of which are toxic and other two non-toxic. Table \ref{tab:corpus} shows size, percentile of toxicity, and the average toxicity of each corpus. 



\section{Experiments}
We conduct adaptive pretraining on four separate GPT-2 models on each corpus discussed in Sec.~\ref{sec:corpus_creation}. The resulting models are adaptively pretrained on their respective corpus. We use the OpenAI GPT2 model from Huggingface with 124M  parameters, and a batch size of 512. We use the Adam optimizer \citep{kingma2014method}, with the learning rate of $5e^{-5}$, and training over three epochs.

\subsection{Decoding with Decay Function}
This step is only required for LMs pretrained on the toxic domain. We first generate a sentence conditioned on the RealToxicityPrompts \cite{Gehman2020REALTOXICITYPROMPTS:Models}. Let $M_{org}$ be an LM that we want to detoxify. In our study, there are two choices for $M_{org}$. One is the default LM without adaptive pretraining. Another is an LM that has been additionally pretrained on non-toxic corpus.  Let $M_{dapt}$ be a language model that has been adaptively pretrained on a toxic corpus. Let \textbf{x} be a prompt that we use to generate continuation. For each consecutive token $w$, we have two probability distributions $p(w\, |\, M_{org}, \textbf{x})$ and  $p(w\, |\, M_{dapt}, \textbf{x})$. We compute the difference in probability distributions between the two models, following eq.~\ref{eq:1}. 
\begin{equation} \label{eq:1}
\Delta p(w, \textbf{x}) = p(w\, |\, M_{org}, \textbf{x}) - p(w\, |\, M_{dapt}, \textbf{x})
\end{equation}

If $p(w, \textbf{x}) < 0$, token $w$ has higher probability of occurring in $M_{dapt}$. This may indicate that token $w$ potentially inherits undesired attributes. We use a scaling function in eq.~\ref{eq:2} to scale down the probability of such words, following \citet{Schick2020Self-DiagnosisNLP}:

\begin{equation} \label{eq:2}
    \alpha(x) = 
\begin{cases}
    1 \hspace{10mm} \text{if}\hspace{3mm}x \geq 0  \\
    e^{\lambda x} \hspace{6mm} \text{otherwise}
\end{cases}
\end{equation}

The hyperparameter $\lambda$ is a decay constant of the scaling function. We set it to 100 as it is proven to reduce the toxicity more effectively than other values \citep{Schick2020Self-DiagnosisNLP}. 

\section{Evaluation}
\subsection{Evaluation on Debiasing} \label{sec:debias_eval}

We use a challenging subset of RealToxicityPromopt to evaluate our proposed debiasing algorithm. Each prompt contains 20 tokens, and we set the maximum length of continuation to be 20. We classify a sentence to exhibit an attribute if the attribute score assigned by the Perspective API is at least 50 \%, following \citet{Gehman2020REALTOXICITYPROMPTS:Models}. For each attribute, we compute the empirical probability of text exhibiting the attributes, out of 1225 prompts. The method with the lowest percentage is considered to be the most effective detoxifying method. 

We compare our approch to the following three baselines:
\begin{itemize}
\setlength{\itemsep}{-5pt}
    \item Default GPT-2,
    \item DAPT on non-toxic corpus, and
    \item Self-debiasing
\end{itemize}
where DAPT on non-toxic corpus represents a data-based approach, and self-debiasing represents a decoding-based approach. We also test the ensemble of existing methods and our proposed method. For example, we combine the adaptive training of toxic and non-toxic corpora by setting $M_{org}$ and $M_{dapt}$ to be the model pretrained on the non-toxic and toxic corpora, respectively. 

\begin{table*}
    \scriptsize
    \centering
    \begin{tabular}{p{2.2cm}rrrrrrrr}
    \hline
 \textbf{Attribute} &  \textbf{Toxicity} & \textbf{Sev. Tox.} & \textbf{Id. Attack} & \textbf{Insult} & \textbf{Threat} & \textbf{Profanity} & \textbf{Sex. Exp.} & \textbf{Flirt.} \\
  \hline
 Default GPT-2 & 38.9&27.4 &11.6 & 31.9 & 16.8 &30.0 &23.9 &27.6\\
 \textbf{$ \;\; + DAPT_{toxic-95}$} &\autour{$\downarrow$ 9.4} 29.5 &\autour{$\downarrow$ 7.7} 19.7 &\autour{$\downarrow$ 3.0} 8.60 &\autour{$\downarrow$ 8.7} 23.2 & \autour{$\downarrow$ 2.0} 14.8& \autour{$\downarrow$ 7.5} 22.5&\autour{$\downarrow$ 4.6} 19.3 &\autour{$\downarrow$ 1.1} 26.5\\
 
 \textbf{$ \;\; + DAPT_{toxic-98}$} & \autour{$\downarrow$ 6.9} 32.0& \autour{$\downarrow$ 6.1} 21.3&\autour{$\downarrow$ 0.8} 10.8 &\autour{$\downarrow$ 6.9} 25.0 & \autour{$\downarrow$ 2.5} 14.3 & \autour{$\downarrow$ 5.1} 24.9 &\autour{$\downarrow$ 3.9} 20.0 &\autour{$\downarrow$ 0.8} 26.8\\
 \arrayrulecolor{gray}\hline
 
 \textbf{$\mathrm{DAPT_{nontoxic-2}}$} &  16.5&  10.2&  5.25&  12.4 & 7.59& 11.8&  9.79&  16.9\\
 
 \textbf{$ \;\; + DAPT_{toxic-95}$} & \autour{$\downarrow$ 7.3} 9.17& \autour{$\downarrow$ 5.8} 4.42&\autour{$\downarrow$ 1.7} 3.59& \autour{$\downarrow$ 5.7} 6.67& \redtour{$\uparrow$ 0.2} 7.76 & \autour{$\downarrow$ 6.0} 5.84&\autour{$\downarrow$ 3.3} 6.42 &\autour{$\downarrow$ 0.9} 16.0\\
 
 \textbf{$ \;\; + DAPT_{toxic-98}$} &\autour{$\downarrow$ 7.7} 8.76 & \autour{$\downarrow$ 5.8} 4.42& \autour{$\downarrow$ 2.1} 3.17& \autour{$\downarrow$ 7.5} 4.92  & \autour{$\downarrow$ 0.3} 7.34& \autour{$\downarrow$ 6.2} 5.59&\autour{$\downarrow$ 3.9}  5.92&\autour{$\downarrow$ 1.4} 15.5\\
 
 \textbf{$\mathrm{DAPT_{nontoxic-5}}$} &  11.2&  6.26 & 3.59 &7.92  &  6.76 & 7.92 & 7.84 & 15.8\\
 
 \textbf{$ \;\; + DAPT_{toxic-95}$} & \autour{$\downarrow$ 5.1} 6.09 & \autour{$\downarrow$ 3.0} 3.25 &\autour{$\downarrow$ 1.1} \textbf{2.50} & \autour{$\downarrow$ 3.7} 4.25 &  \autour{$\downarrow$ 1.6} 5.17 & \autour{$\downarrow$ 4.0} 3.92 &\autour{$\downarrow$ 3.2} \textbf{4.67}&\autour{$\downarrow$ 4.6} \textbf{11.2}\\
 
 \textbf{$ \;\; + DAPT_{toxic-98}$} &\autour{$\downarrow$ 5.5} \textbf{5.75} & \autour{$\downarrow$ 3.8} \textbf{2.50} &\autour{$\downarrow$ 0.8} 2.75 & \autour{$\downarrow$ 4.5} \textbf{3.42}  &\autour{$\downarrow$ 1.7} \textbf{5.09} &\autour{$\downarrow$ 4.3} \textbf{3.59} & \autour{$\downarrow$ 2.7} 5.17 & \autour{$\downarrow$ 3.4} 12.4\\
 
 \hline
 Self-Debiasing & 31.7& 21.2&10.0 & 24.0 & 15.0& 23.9& 17.3&24.4\\
    \end{tabular}
    \caption{Empirical probabilities of the eight attributes on RealToxicityPrompts.}
    \label{tab:result}
\end{table*}

\begin{table}
    \centering
    \begin{tabular}{l|rr}
        \textbf{Domain} &  \textbf{Default} & \textbf{Debiasing} \\ \hline
        \textbf{American Actor} & 2.94 &  \autour{$\downarrow$ 2.33} 0.61\\
        \textbf{American Actress} & 4.07 & \autour{$\downarrow$ 3.81} 0.26\\ \
        \textbf{Left} & 8.47 & \autour{$\downarrow$ 8.47} 0.00 \\
        \textbf{Right} & 5.08 & \autour{$\downarrow$ 5.08} 0.00 \\ 
        \textbf{Asian} & 1.94  &  \autour{$\downarrow$ 1.94} 0.00\\
        \textbf{African} & 5.83 & \autour{$\downarrow$ 5.83} 0.00\\
        \textbf{European} & 5.83&\autour{$\downarrow$ 2.92 } 2.91 \\
        \textbf{Hispanic/Latino} & 2.91 & \autour{$\downarrow$ 0.97} 1.94\\
    \end{tabular}
    \caption{Empirical probabilities of the Toxicity attribute on BOLD. The Debiasing method is $ DAPT_{toxic-5} + DAPT_{toxic-98}$.}
    \label{tab:bold_result}
\end{table}

\section{Results and Discussion}

Table \ref{tab:result} shows the empirical probability of generating text exhibiting an attribute, conditioned on the challenging prompts of the RealToxicityPrompts dataset. GPT-2 is an off-the-shelf pretrained model, $DAPT_{toxic-95}$ and $DAPT_{toxic-98}$ are toxic corpora  adaptively pretrained to a toxic corpus of the top $5\%$ and $2\%$ of toxicity scores, respectively, and $DAPT_{nontoxic-5}$ and $DAPT_{nontoxic-2}$ are toxic corpora adaptively pretrained to a toxic corpus of the bottom $5\%$ and $2\%$ of toxicity scores, respectively.

\subsection{Data-based over Decoding-based}
 Without debiasing, the probability of generating text exhibiting toxicity approaches 40\%. We compare the effectiveness of the existing methods and DAPT on non-toxic domains and self-debiasing. 
 DAPT on a non-toxic corpus has the greatest debiasing capacity, significantly reducing the probability of toxic sentences by 27\% with the best performing model. 
 
 \begin{figure}[htp]
    \centering
    \includegraphics[width=8cm]{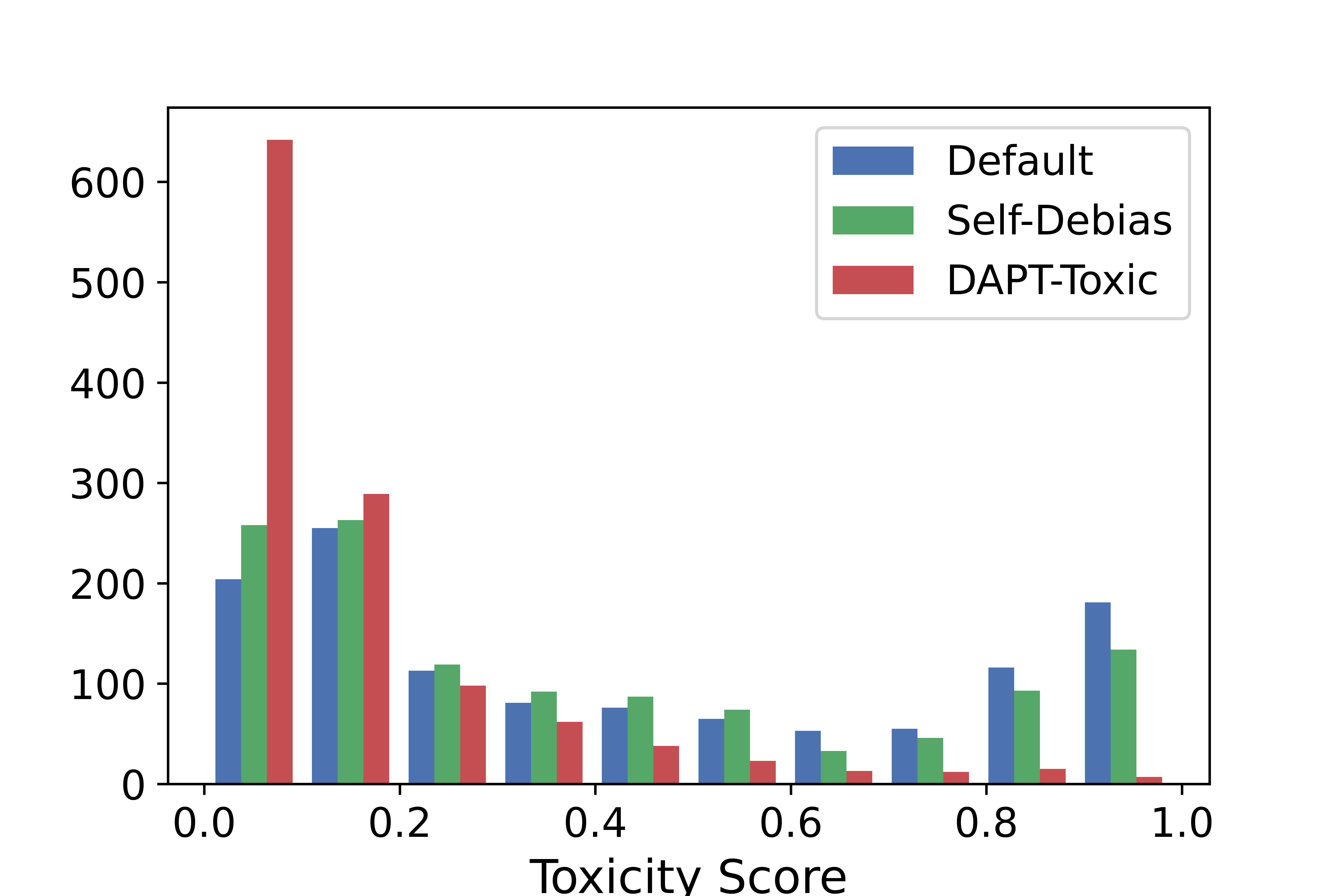}
    \caption{The distribution of toxicity scores conditioned on the challenging subset of RealToxicityPrompts.}
    \label{fig:toxicity_plot}
\end{figure}

\subsection{Toxic Corpora Help Reduce Toxicity}
When combining the existing method with our proposed method, the empirical probability is reduced with varying degrees, indicating the complementary effect of the toxic corpus. Table \ref{tab:result} shows that the most effective debiasing approach is $DAPT_{nontoxic-5}$ + $DAPT_{toxic-98}$ and  $DAPT_{nontoxic-5}$ + $DAPT_{toxic-95}$, each achieving the best score on different attributes. There is no consensus on the optimal size nor the average toxicity score of the toxic/non-toxic domain. This might depend on the objective of a task. 

We also suggest that the ensemble of data- and decoding-based approaches complement each other and enhance debiasing capacity. In Figure \ref{fig:toxicity_plot}, our proposed method $DAPT_{nontoxic-5}$ + $DAPT_{toxic-98}$ produces approximately 80 \% of sentences in the range between 0.00 and 0.20, showing the most significant effectiveness. 

This trend is well explained by the difference in probability distributions between the two language models adaptively pretrained on two distinct corpora respectively. Since  $DAPT_{toxic-98}$ tends to produce toxic context with higher probabilities, there is a higher chance of being penalized by the decay function (eq.~\ref{eq:2}).

\section{Conclusion}
Large pretrained LMs suffer from degeneration and exhibit biases and toxicity despite their vast capabilities. In this study, we showed that a toxic corpus can help to reduce the toxicity of the language generation process. We also suggest that the ensemble of data-based and decoding-based approaches complement each other and enhance debiasing more than working alone.

\section*{Acknowledgments}
We would like to acknowledge the Vector Institute of Artificial Intelligence for providing computing resources. This research is funded by a Vector Institute Research Grant. Rudzicz is supported by a CIFAR Chair in AI.

\bibliography{anthology, references}
\bibliographystyle{acl_natbib}





\end{document}